\documentclass{article}
\usepackage{ijcai26}

\usepackage{times}
\usepackage{soul}
\usepackage{url}
\usepackage[hidelinks]{hyperref}
\usepackage[utf8]{inputenc}
\usepackage[small]{caption}
\usepackage{graphicx}
\usepackage{amsmath}
\usepackage{amsthm}
\usepackage{booktabs}
\usepackage{algorithm}
\usepackage{algorithmic}
\usepackage[switch]{lineno}
\usepackage{multirow}
\usepackage{makecell}
\usepackage{pifont}
\usepackage{xcolor}

\newcommand{\q}[1]{$\mathbf{Q#1}$}

 \author{
 William Solow$^1$
 \and
 Paola Pesantez-Cabrera$^2$
 \and
 Markus Keller$^2$
 \and 
 Lav Khot$^2$
 \and
 Sandhya Saisubramanian$^1$\and
 Alan Fern$^1$
 \affiliations
 $^1$Oregon State University\\
 $^2$Washington State University\\
 \emails
 \{soloww, sandhya.sai, afern\}@oregonstate.edu, \{p.pesantezcabrera, mkeller, lav.khot\}@wsu.edu
 }

\title{A Hybrid Modeling Framework for Crop Prediction Tasks via Dynamic Parameter Calibration and Multi-Task Learning}

\begin{document}

\maketitle

\begin{abstract}
Accurate prediction of crop states (e.g., phenology stages and cold hardiness) is essential for timely farm management decisions such as irrigation, fertilization, and canopy management to optimize crop yield and quality. While traditional biophysical models can be used for season-long predictions, they lack the precision required for site-specific management. Deep learning methods are a compelling alternative, but can produce biologically unrealistic predictions and require large-scale data. We propose a \emph{hybrid modeling} approach that uses a neural network to parameterize a differentiable biophysical model and leverages multi-task learning for efficient data sharing across crop cultivars in data limited settings. By predicting the \emph{parameters} of the biophysical model, our approach improves the prediction accuracy while preserving biological realism. Empirical evaluation using real-world and synthetic datasets demonstrates that our method improves prediction accuracy by 60\% for phenology and 40\% for cold hardiness compared to deployed biophysical models. Project site with additional details: \url{https://tinyurl.com/DMC-MTL-Site}.
\end{abstract}

\begin{figure*}[t]
    \centering
    \includegraphics[width=0.9\linewidth]{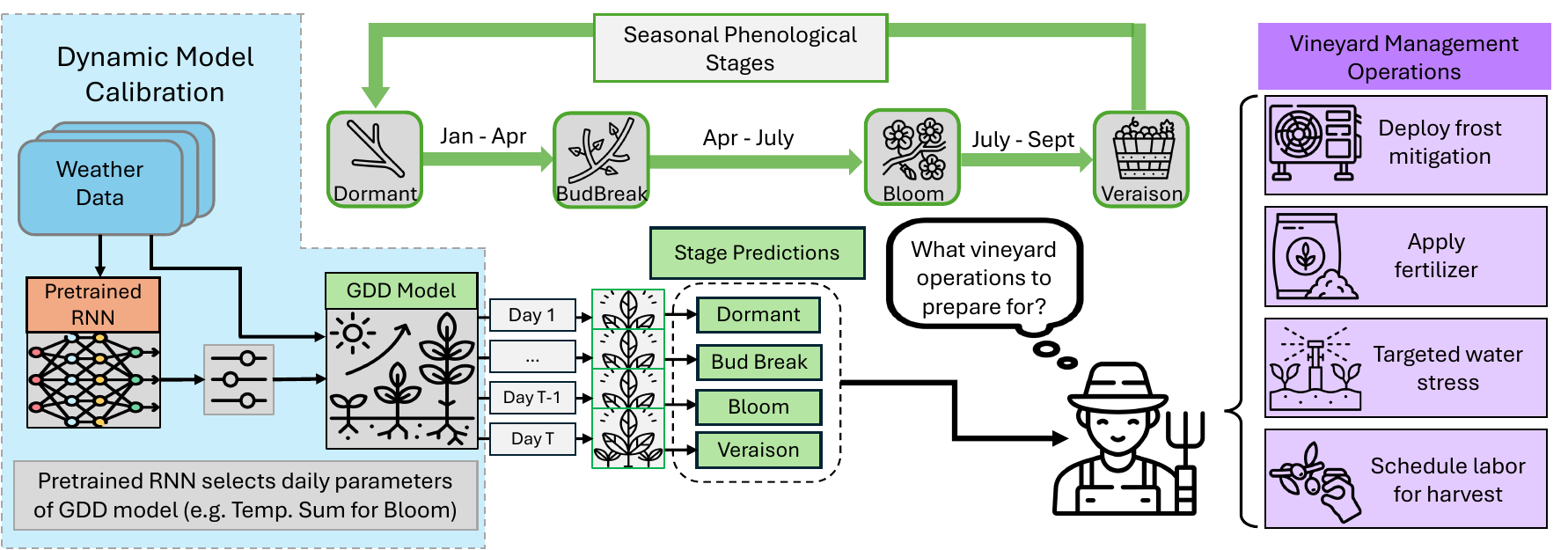}
    \caption{Overview of our proposed method using phenology prediction as an example. In this case, seasonal phenological stages guide vineyard management operations. Our Dynamic Model Calibration via Multi-Task Learning (DMC-MTL) approach uses a pretrained neural network and biophsyical model to produce high fidelity and biologically realistic phenology state forecasts. The pretrained RNN produces daily parameter predictions (Base Temperature, Temperature Sum for Bud Break, etc.) of the Growing Degree Day (GDD) phenology model which handles the daily stage prediction.}
    \label{fig:statement_of_problem}
    \vspace{-5pt}
\end{figure*}

\section{Introduction}
Accurate forecasts of crop states are critical for growers to schedule time-sensitive farm operations such as cold stress mitigation, pruning, fertilization, irrigation, and harvesting for specialty crops~\cite{keller2016,milani2024,rogiers2022}. However, accurate crop state prediction is challenging due to (1) limited availability of historical data for per-cultivar calibration and sparse observations during each growing season~\cite{zapata2017}, and (2) the need to accurately model complex relationships between daily weather features and crop states~\cite{guralnick2024a}. Existing approaches to this crop state prediction problem typically fall into two categories: mechanistic biophysical models and data-driven deep learning approaches.

Historically, biophysical models have been used to model a variety of crop states, such as phenology (timing of each developmental stage) and cold-hardiness (tolerance to low temperatures). Phenology is modeled by the Growing Degree Day (GDD) biophysical model based on daily accumulated heat units~\cite{parker2013}. Cold hardiness models predict the lethal bud temperature as a function of air temperature and phenological stage~\cite{ferguson2011}. Despite research supporting that crop states depend on exogenous weather features~\cite{greer2006}, most specialty crop models only use air temperature as input, limiting their expressiveness~\cite{badeck2004}. Further, current biophysical models do not capture temporal nuances, such as how  varied chilling hours in winter can change dormancy release and phenological development~\cite{keller2010}. These limitations affect their ability to produce accurate medium range (7-14 day) forecasts of the crop state~\cite{reynolds2022}. 

Deep learning offers a compelling alternative to biophysical modeling due to its ability to model complex, nonlinear, and temporal relationships between weather variables and crop physiological states. However, purely data-driven models typically require large datasets and often produce \emph{biologically unrealistic} predictions that violate known physiological constraints on plant development, such as predicting bud break after flowering. This makes them unsuitable for actionable medium range forecasts~\cite{saxena2023b}. 

To address the limitations of both deep learning and biophysical modeling for crop state forecasting tasks, we present a hybrid approach that uses a recurrent network to \emph{dynamically refine the parameters} of a differentiable biophysical crop model, based on daily weather (Figure~\ref{fig:statement_of_problem}). 
For example, given historical weather data and phenology observations, the network is trained to assign the base temperature for heat accumulation in the GDD model such that the calibrated GDD model can accurately predict various phenological stages at deployment time. To address limited per-cultivar data, we use multi-task learning~\cite{caruana1997} to share information across cultivars and improve prediction accuracy.

While hybrid modeling techniques have been successfully applied to a variety of physical processes~\cite{jia2021,cai2021}, they remain underexplored for crop state tasks. Prior hybrid methods often approximate portions of the biophysical model~\cite{bree2025} or predict residuals~\cite{vijayshankar2021}. In contrast, our approach predicts parameters of the biophysical model, thereby producing biologically realistic and more accurate predictions. 

\paragraph{Scope, Contributions, and Track Relevance} This work is motivated by the operational needs of our primary stakeholders---specialty crop growers in the Pacific Northwest (PNW) of the United States who face increasing risks from weather variability and climate stress~\cite{reynolds2022}. In this region, over \$10 billion in annual specialty-crop production depends on timely and accurate crop state forecasts, including phenology and cold hardiness~\cite{knowling2021}. Our team includes AI experts and agricultural domain experts who played key roles in data collection, selection of biophysical models, and system deployment. 

While our framework is broadly applicable to other crops, the paper focuses on wine grapes as a representative specialty crop and studies two critical crop states: \emph{phenology} and \emph{cold hardiness}. We train and validate our models using grapevine data collected in the PNW region between 1988-2025. 

Our primary contributions are: (1) presenting a novel hybrid approach for accurate crop state forecasting by refining parameters of the biophysical model, conditioned on the weather features; (2) formulating the crop state prediction problem as a multi-task learning problem that leverages data efficiently across grape cultivars; (3) presenting an in-season adaptation variant of our model; and (4) empirical evaluations using real-world and synthetic datasets that demonstrate our approach's robustness and increased accuracy over state-of-the-art biophysical baselines, deep learning approaches, and hybrid models. Our model has been \emph{deployed} on AgWeatherNet~\cite{c:agnet} which has over 26000 registered growers. 

\section{Background and Related Work}

\paragraph{Hybrid Modeling of Biophysical Processes}
Hybrid modeling combines deep learning and mechanistic modeling to obtain increasingly accurate and interpretable models of biophysical processes~\cite{willard2023}. Physics-Informed Neural Networks (PINNs; \cite{raissi2019}) encode biological constraints in the form of partial differential equations into neural networks, enabling more accurate and biologically realistic predictions~\cite{karpatne2017a,karniadakis2021}. In addition to PINNs, residual error hybrid models use a deep learning model to predict the difference between the true observation and the biophysical model prediction~\cite{demattosneto2022}. Process replacement hybrid models~\cite{wang2023a} replace a poorly understood aspects of the biophysical model with a neural network with wide applications to the natural sciences~\cite{feng2022,shen2023}. Table~\ref{tab:modeling_options} summarizes the characteristics of these models along different desiderate for crop state forecasting tasks. 
Another related work is that of \citeauthor{unagar2021}~(\citeyear{unagar2021}) that use reinforcement learning for parameter calibration of a lithium battery. However, their problem setting assumes that the next true state of the system is known, which is untrue in our problem setting where medium range forecasts are needed. We adopt a supervised learning approach and train a network to dynamically modulate the parameters of the biophysical model in response to input features, enabling medium horizon forecasts. 

\paragraph{Multi-Task Learning}
When a set of tasks share similar structure, the multi-task learning \cite{zhang2018a} framework can be used to efficiently aggregate data across tasks. Hard parameter sharing methods share a common set of hidden layers with unique prediction heads while soft parameter sharing methods learn unique models but regularize weight updates to keep models similar~\cite{ruder2017}. Shared embedding spaces~\cite{caruana1997} and task-specific embeddings~\cite{changpinyo2018} are viable methods for encoding task-specific information. We empirically evaluate these approaches in data-limited setting across genetically diverse grape cultivars treated as separate tasks, assessing prediction accuracy for all cultivars.

\begin{table}[t]
    \centering
    \resizebox{\columnwidth}{!}{%
    \begin{tabular}{l|ccc}
    \toprule
         Modeling Approach & \makecell{Biologically\\Realistic} & \makecell{Exogenous\\Features} & \makecell{Temporal\\Info.}\\
         \midrule
         \midrule
         Biophysical Model  & \textcolor{green}{\ding{51}} & \textcolor{red}{\ding{55}} & \textcolor{red}{\ding{55}} \\
         Deep Learning  & \textcolor{red}{\ding{55}} & \textcolor{green}{\ding{51}} & \textcolor{green}{\ding{51}} \\
         PINN  & Sometimes & \textcolor{green}{\ding{51}} & \textcolor{green}{\ding{51}} \\
         Residual & Sometimes & \textcolor{green}{\ding{51}} & \textcolor{green}{\ding{51}} \\
         Process Replacement  & \textcolor{green}{\ding{51}} & \textcolor{red}{\ding{55}} & \textcolor{red}{\ding{55}} \\
         \textbf{DMC-MTL (Ours)}& \textcolor{green}{\ding{51}} & \textcolor{green}{\ding{51}} & \textcolor{green}{\ding{51}} \\
         \bottomrule
    \end{tabular}
    }
    \caption{Modeling approaches for crop state forecasting tasks evaluated along different desiderata. PINN, Residual, Process Replacement, and DMC-MTL are all hybrid models. \emph{Biologically realistic} means the model predictions reflect biological laws. \emph{Exogenous features} means that the model output can be conditioned on additional weather features. \emph{Temporal info.} indicates that the model can leverage historical input in addition to what the biophysical model uses.}
    \label{tab:modeling_options}
    
\end{table}

\paragraph{Machine Learning for Crop State Forecasts}
\citeauthor{saxena2023b}~(\citeyear{saxena2023b}) applied multi-task learning to grape bud break prediction using a classification model. However, this model made erroneous predictions (e.g., predicting the onset of dormancy after bud break) that were inconsistent with biological processes. \citeauthor{saxena2023a}~(\citeyear{saxena2023a}) framed the grape cold hardiness prediction problem as a multi-task learning problem and used a recurrent neural network (RNN) to improve prediction accuracy over the deployed Ferguson biophysical model for cold hardiness~\cite{ferguson2014}, demonstrating efficacy of multi-task learning to leverage data across cultivars. \citeauthor{bree2025}~(\citeyear{bree2025}) proposed a process replacement hybrid model for bloom date in cherry trees by approximating the temperature response function in the GDD model with a neural network. However, their method did not consider the effect of exogenous weather features on phenology nor the temporal variation of heat accumulation. In contrast, our method leverages a RNN to both encode temporal and exogenous weather information to avoid these limitations, and demonstrates state of the art performance in predicting grape phenology and matches grape cold hardiness while producing biologically realistic forecasts. 

\section{Dynamic Model Calibration}

Our problem setting and approach, \textbf{D}ynamic \textbf{M}odel \textbf{C}alibration with \textbf{M}ulti-\textbf{T}ask \textbf{L}earning (DMC-MTL), is inspired by the following observations and hypotheses: (1) crop state prediction tasks are similar across cultivars, motivating our multi-task approach; (2) the parameters of crop state prediction models are hypothesized to vary over time based on historical weather in addition to the daily average temperature; and (3) crop state forecasting requires biologically realistic and accurate predictions. 

\begin{figure}[t]
    \centering
    \includegraphics[width=0.9\linewidth]{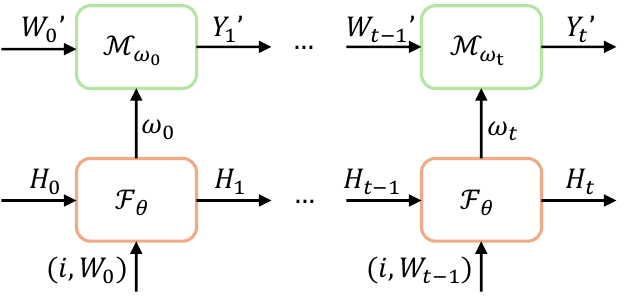}
    \caption{The network architecture of our approach DMC-MTL. The multi-task RNN ($\mathcal{F}_\theta$) sequentially embeds cultivar i.d. $i$ and concatenates it with the daily weather features $W_t$ to predict a parameterization $\omega_t$ of the biophysical model $\mathcal{M}$. Using the weather input to the biophysical model $W_t'$ and the daily parameterization $\omega_t$, crop state forecasts $Y_t'\ldots Y_{t+k}'$ can be generated.}
    \label{fig:model_arch}
\end{figure}

\paragraph{Problem Formulation}
We formulate the problem of estimating dynamic parameters of a biophysical model as a time series supervised learning problem and adopt the multi-task setting. Let $\mathcal{M}_\omega$ denote the biophysical model with parameters $\omega$. Let $D_i$ be the set of observed weather and daily crop states for each crop cultivar $i$. Let $S_{i,k}$ be the $k$-th season in $D_i$ with $S_{i,k}\!=\!\{W_0,Y_0,\ldots,W_T,Y_T\}$ where $W_t$ is the observed weather feature vector and $Y_t$ is the observed crop state on day $t$. Given $W_t'\!\subset\!W_t$ as input (the biophysical model input is lower dimensional than the total number of observed weather features), $\mathcal{M}_\omega$ predicts a crop state $Y_t'$.
We train a multi-task recurrent neural network model $\mathcal{F}_\theta$ that takes $W_t$ and cultivar ID $i$ as input, and outputs daily parameters $\omega_t$ of $\mathcal{M}$. The resulting parameterized model $\mathcal{M}_{\omega_t}$, along with $W_t'$, is used to generate crop state predictions $Y_t'$. Given time series input $S_{i,k}$, we use $\mathcal{F}_\theta$ and $\mathcal{M}$ to obtain a sequence of parameter estimates $\omega_0,\ldots\omega_T$ and corresponding crop state predictions $Y_0',\ldots,Y_T'$ (Figure~\ref{fig:model_arch}).

\subsection{Model Architecture}
The proposed model architecture for DMC-MTL is comprised of three parts: the RNN-backbone, the multi-task model, and the parameterization of the biophysical model. 

The RNN-backbone ($f_\theta$) contains two linear layers, followed by a Gated Recurrent Unit (GRU;~\cite{chung2014}), and another linear layer. To support multi-task learning across cultivars, we define $\mathcal{F}_\theta$ which adds a linear embedding layer before $f_\theta$. This embedding layer converts a one-hot encoding of the cultivar into a dense vector, which is concatenated with the daily weather feature vector $W_t$ and passed to $f_\theta$, allowing the model to incorporate cultivar-specific information~\cite{saxena2023a}. ReLU activations are used, except for the final layer where a $\tanh$ activation is applied. The output of $\mathcal{F}_\theta$, which is in the range $[-1, 1]$, is then rescaled to match the parameter ranges of the biophysical model $\mathcal{M}$. 

Figure~\ref{fig:model_arch} shows the daily parameterization $\omega_t$ of the biophysical model $\mathcal{M}$, the core of our DMC-MTL approach. $\mathcal{F}_\theta$ makes causal parameter predictions by sequentially processing a weather data sequence $W_0,\ldots, W_T$, generating corresponding parameters $\omega_t$ at each time step. These parameters are used to parameterize $\mathcal{M}_{\omega_t}$ and along with $W_t'$, to produce phenology prediction $Y_{t+1}'$. 

\paragraph{Biophysical Model Implementation} 
To learn $\mathcal{F}_\theta$, the biophysical model $\mathcal{M}$ must be differentiable and implemented in a framework that supports gradient backpropagation. In practice, most specialty crop state models are relatively simple and do not require advanced ordinary differential equation solvers. We additionally modify each biophysical model so that the parameters can be updated daily by $\mathcal{F}_\theta$ before each integration step. Parameters for the biophysical models are known to lie in specified ranges. To retain the ability of the DMC-MTL approach to capture complex dependencies among weather features, we choose large ranges for each parameter. Parameters have biophysical meaning; the Base Temp for Emergence parameter is the minimum temperature required before the grapevine can exit dormancy.

\section{Extending DMC-MTL: In-Season Tuning}
Collecting sufficient historical data for training accurate DMC-MTL models is challenging and often requires many years. In the case of insufficient data, it is beneficial to train the DMC-MTL model with available data and then adjust the model parameters based on \emph{in-season observations} made by growers in the field. To enable in-season adaptation, \emph{error signals} are generated by comparing DMC-MTL model predictions with sparse in-season observations such as occasional measurements of bud stage or cold hardiness. These error signals are then used to adjust future predictions of biophysical model parameters within the same growing season. This continual recalibration reduces bias accumulation seen in static prediction models and improves prediction accuracy. 

\begin{figure}[t]
    \centering
    \includegraphics[width=0.9\linewidth]{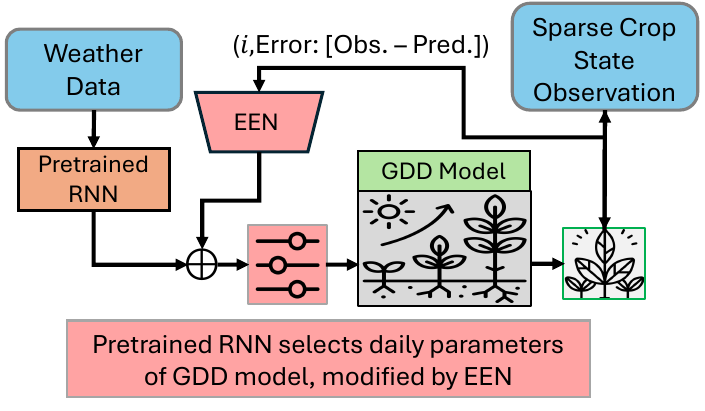}
    \caption{In-Season Adaptation with DMC-MTL. Cultivar i.d. $i$ and error between observed and DMC-MTL predicted crop states are passed to the EEN. The parameters predicted by the EEN are combined additively to the prediction made by DMC-MTL's pretrained RNN before parameterizing the biophysical model. }
    \label{fig:een_hybrid}
\end{figure}

When in-season observations are available, the daily parameter predictions made by the DMC-MTL approach are adjusted by using an additional neural network conditioned on error signals (Figure~\ref{fig:een_hybrid}). While such Error Encoding Networks (EENs) have been used for video frame prediction~\cite{henaff2017}, they have not been explored for the \emph{sparse and low-dimensional} error signals available in crop state tasks. Furthermore, EENs have not previously been applied to hybrid models, so we investigate their effectiveness in reducing in-season predictive error and recalibrating the hybrid DMC-MTL model at small-to-medium horizon lengths.

\paragraph{Model Architecture}
We use an EEN with the same model architecture as DMC-MTL ($\mathcal{F}_\theta$) given that observations are sparse and different cultivars may exhibit different error patterns. At each time step $t$ the DMC-MTL model makes a prediction. If an in-season observation is available, the difference between the observation $Y_t$ and model prediction $Y_t'$ is passed to the EEN. If no observation is available, the EEN input is zero. This is consistent with the observation that if the DMC-MTL model makes no error, then the EEN should not change the future model predictions of the biophysical model parameters. We  accomplish this by setting the bias terms within the EEN network layers to zero. EEN parameter predictions are combined \emph{additively} with the predicted parameters by the pretrained DMC-MTL and then passed to the biophysical model to predict the next crop state.    

\paragraph{Training EENs} Training the In-Season Adaptation model is a two step process. First, a DMC-MTL model is trained to predict parameters of the biophysical model that best predict the observed crop state. Then, the weights of the pretrained RNN $\mathcal{F}_\theta$ are frozen while the EEN is trained. The EEN is trained using the same training data as DMC-MTL, under the assumption that part of the observed crop state cannot be predicted solely by weather and cultivar ID~\cite{henaff2017}. We use historical crop state observations to mimic the availability of in-season observations in real time. We randomly mask available observations to ensure that the EEN does not learn rely on frequent observations for medium range forecasts, which may not be available at deployment time.

\begin{table*}[t]
\centering
\setlength{\tabcolsep}{4mm}
{\fontsize{9}{09}\selectfont
\begin{tabular}{llrr}
\toprule
&      Solution Approaches             & Grape Phenology & Grape Cold Hardiness  \\
\midrule
\midrule
\multirow{7}{*}{\textbf{Q1a:}} & 
                      \bf{DMC-MTL (Ours)}   & \bf{7.63 $\pm$ 3.56}\phantom{$^*$}   & \bf{1.21 $\pm$ 0.39}\phantom{$^*$}  \\
                      & Bio. Model (Deployed)      & 18.58 $\pm$  5.03$^*$  & 2.03 $\pm$ 0.39$^*$   \\
                      & Bio. Model (Gradient Descent)      & 12.21 $\pm$  5.13$^*$ & 1.88 $\pm$ 0.42$^*$    \\
                     & Deep-MTL              & 8.16  $\pm$  4.20$^*$                & 1.30 $\pm$ 0.46\phantom{$^*$}     \\
                     & TempHybrid            & 9.84 $\pm$   4.35$^*$                & 3.45 $\pm$ 0.98$^*$               \\
                     & PINN                  & 8.61  $\pm$  4.32$^*$                & 1.30 $\pm$ 0.43\phantom{$^*$}     \\
                     & Residual Hybrid       & 15.01 $\pm$  6.00$^*$                & 1.49 $\pm$ 0.52$^*$               \\ 
                     \midrule
\multirow{5}{*}{\textbf{Q2:}}  
                     & DMC-STL          & 9.57  $\pm$ 3.79$^*$   & 1.62 $\pm$ 0.34$^*$  \\
                     & DMC-Agg          & 9.81  $\pm$ 4.70$^*$   & 1.51 $\pm$ 0.70$^*$  \\
                     & DMC-MTL-Mult     & 11.97 $\pm$ 4.43$^*$   & 1.57 $\pm$ 0.53$^*$  \\       
                     & DMC-MTL-Add      & 8.42  $\pm$ 3.56$^*$   & 1.26 $\pm$ 0.41\phantom{$^*$}  \\       
                     & DMC-MTL-MultiH   & 8.20  $\pm$ 4.15$^*$   & 1.34 $\pm$ 0.48$^*$  \\  
\bottomrule
\end{tabular}
}
\caption{The average seasonal error (RMSE in days for phenology and $^\circ C$ for cold hardiness) over all cultivars and five seeds in the testing set for grape phenology and cold hardiness. DMC-MTL is compared against two optimization procedures for the biophysical model, a deep learning approach, three hybrid models, and two DMC variants. Best-in-class results are reported in bold. A $^*$ indicates that DMC-MTL yields a \textit{statistically significant improvement} ($p < 0.05$) using the paired t-test relative to the corresponding baseline.}
\label{tab:main_results}
\end{table*}

\section{Experiment Setup}
The performance of our proposed DMC-MTL approach is evaluated on two key criteria: (1) accurate and biologically realistic seasonal predictions and (2) efficient data use across cultivars. We also considered (3) how in-season data can be used to reduce prediction error and (4) robustness to unexpected subseasonal weather patterns, which can be demonstrated by evaluating a model on different weather distribution. Based on these, we design our experiments to answer the following research questions: 
\begin{enumerate}
\setlength{\itemsep}{0cm}
    \item[\q{1}:] {(a) How does the average seasonal accuracy of DMC-MTL compare to deployed biophysical, deep learning and hybrid models? (b) Are predictions biologically realistic?} 
    \item[\q{2}:] {(a) Does DMC-MTL leverage data efficiently across cultivars? (b) How much per-cultivar data is required?}
    \item[\q{3}:]{Do in-season crop state observations improve DMC-MTL model predictions?}
    \item[\q{4}:] {Does DMC-MTL exhibit robustness to different weather conditions compared to other baselines?}
    \item[\q{5}:] {What percentage of cultivars are accurately predicted by each model type? }
\end{enumerate}

\subsection{Datasets} 
\paragraph{Real-World Datasets} We use the grape phenology and cold hardiness of 32 grape cultivars collected between 1988-2025. Phenology (stages of bud break, bloom, and veraison) was observed daily during the non-dormant season, and cold hardiness was measured weekly, biweekly, or monthly during the dormancy season. There are between eight and 21 years of phenological data per cultivar, and between four and 27 years of cold hardiness data per cultivar (43 to 797 samples). Pertinent historical open field weather data is sourced from AgWeatherNet~\cite{c:agnet}. 

\paragraph{Synthetic Datasets} To explore the robustness of DMC-MTL to different weather conditions, we generated datasets with two biophysical crop models: (1) the \emph{GDD} model with 31 cultivars~\cite{solow2025}, and (2) the \emph{Ferguson} cold hardiness model with 20 cultivars~\cite{ferguson2011}. We used historical weather data from the NASAPower database for Washington, USA~\cite{c:nasapower}. We also generated phenology and cold hardiness observations from Vermont, California, and Oregon, USA. For each biophysical model, we generated ten years of data per cultivar using the biophysical models and historical NASA weather data. We randomly masked $88\%$ of the daily cold hardiness samples to resemble the real-world dataset. 

\subsection{Baselines, Training and Evaluation}

\paragraph{Baselines} We consider 11 baselines for our experiments: (1) \emph{Deployed biophysical model}---the \emph{GDD} model for phenology and the \emph{Ferguson} model for cold hardiness that are used by our stakeholders; (2) \emph{Gradient Descent} on the biophysical model parameters. To the best of our knowledge, this baseline has not been used before because the crop models have not been written in a differentiable framework; (3) \emph{TempHybrid}---a hybrid model proposed by~\citeauthor{bree2025}~(\citeyear{bree2025}) and adapted to the cold hardiness setting; (4) \emph{Deep-MTL}---a multi-task model that either predicts probabilities for each phenological stage, or a continuous approximation of the cold hardiness. Instead of a $\tanh$ activation, there was a single output feature with no activation function. For cold hardiness we used the regression model proposed by~\citeauthor{saxena2023a}~(\citeyear{saxena2023a}); (5) \emph{PINN}---a Physics-informed neural network (PINN) with the same architecture and activation as the Deep-MTL model, trained with an additional loss term to weight biologically realistic predictions~\cite{aawar2025}; (6) \emph{Residual Hybrid}---a hybrid model that uses an RNN to predict the difference between the biophysical model predictions and the observed crop state with the same network architecture as the Deep-MTL model~\cite{vijayshankar2021}; (7) \emph{DMC-STL}---a DMC model without the embedding layer, trained on a per-cultivar basis; (8) \emph{DMC-Agg}---a DMC model without the embedding layer and trained on all unlabeled cultivar data; (9) \emph{DMC-MTL-Mult}---a DMC-MTL variant that uses a multiplicative embedding; (10) \emph{DMC-MTL-Add}---a DMC-MTL variant that uses an additive embedding; (11) \emph{DMC-MTL-MultiH}---a DMC-MTL variant that uses per-task prediction heads.

Baselines 1-6 to evaluate the efficacy of our approach against the biophysical, hybrid and deep learning baselines. Baselines 7-11 evaluate the efficacy of multi-task learning. 

\paragraph{Model Training Protocol} For all experiments, we split the available grape cultivar data into training and testing sets. To build the test set, we withheld two seasons of data per cultivar from the training set. For the cultivars with the least amount of data, this resulted in two years of data in both the training and testing sets. Hyperparameters were selected using a validation set consisting of one season per cultivar. 

Every model was trained for 400 epochs using a learning rate of 0.0002. We decreased the learning rate by a factor of 0.9 after a 10 epoch plateau of the training loss. For the deep learning phenology model we used Cross Entropy loss and used PINN loss for the PINN hybrid model with $p=0.5$. For all other models, we used mean squared error loss function, masking days that did not have a ground truth observation. 

\paragraph{Evaluation Protocol}
We trained each model five times with different data splits and reported the average root mean squared error (RMSE) across cultivars on the test sets. For phenology, the RMSE was the cumulative error in days over the predictions for bud break, bloom, and veraison. For cold hardiness we reported the RMSE in degrees Celsius over all unmasked samples during the testing year. 

\section{Results and Discussion}

\paragraph{Q1a: Average Performance of DMC-MTL}
Table~\ref{tab:main_results} shows the average RMSE values for grape phenology and cold hardiness predictions using different approaches, across 32 cultivars. The results show that DMC-MTL dramatically outperformed the biophysical models that are currently used by growers in PNW region---the GDD model for phenology and Ferguson model for cold hardiness. DMC-MTL also improved over gradient descent optimization for both phenology and cold hardiness. These results indicate the importance of better optimization procedures for crop model calibration, and how DMC-MTL can improve upon standard practices. Further, DMC-MTL improved upon the Deep-MTL, TempHybrid, PINN, and Residual Hybrid models, demonstrating the importance of dynamic parameterization, inclusion of exogenous weather features, and temporal information. We performed the paired t-test aggregated over all cultivars to confirm when our DMC-MTL performance improvements were statistically significant $(p<0.05)$. Overall, our results indicate that our hybrid modeling approach is more accurate and reliable for predicting crop state tasks.

\begin{figure}[t]
    \centering
    \includegraphics[width=0.9\linewidth]{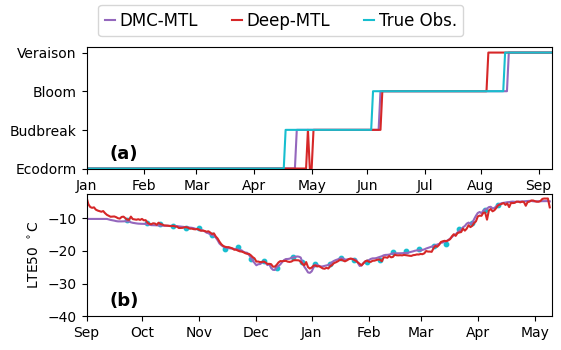}
    \caption{DMC-MTL, Classification, and Regression model predictions for (a) grape phenology and (b) grape cold hardiness. DMC-MTL makes biologically realistic predictions while deep learning model predictions do not always respect biologicaly laws. }
    \label{fig:model_predictions}
\end{figure}

\paragraph{Q1b: Biological Realism} Biologically realistic predictions are critical for interpreting medium-range forecasts that growers rely on to plan their vineyard operations. Figure~\ref{fig:model_predictions} shows that Deep-MTL model incorrectly predicts bloom, reverts to bud break, then re-enters bloom three days later. Similarly, for cold hardiness, Deep-MTL overestimates the biologically plausible cold hardiness early in the growing season. In contrast, our DMC-MTL consistently makes biologically realistic crop state predictions.

\paragraph{Q2a: Multi-Task Data Sharing} 
Results in Table~\ref{tab:main_results} show that DMC-MTL outperforms DMC-STL and DMC-Agg, demonstrating that both single-task learning and naive aggregation are insufficient for these crop state prediction tasks. Other implementations of multi-task learning (DMC-MTL-Mult/Add/MultiH) exhibited lower prediction accuracy compared to DMC-MTL. These results demonstrate that our learned one-hot concatenation embedding approach best encoded task-specific information, thereby efficiently leveraging data between cultivars. 

\begin{figure}[t]
    \centering
    \includegraphics[width=0.9\linewidth]{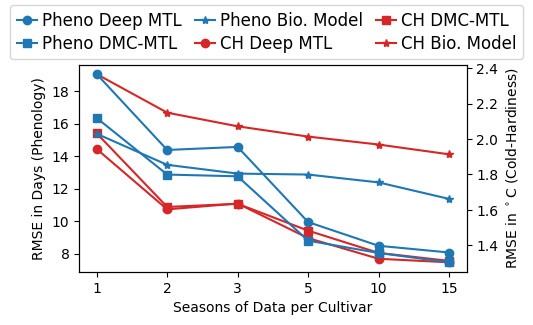}
    \caption{The performance of DMC-MTL models compared to deep learning and biophysical models under limited per-cultivar training data for grape phenology (Pheno) and cold hardiness (CH). Results are averaged over five seeds using the same two-seasons-per-cultivar evaluation sets.}
    \label{fig:data_lim}
\end{figure}

\paragraph{Q2b: DMC-MTL Data Requirements}
To evaluate the per-cultivar data requirements of DMC-MTL and compare its data efficiency to other models, we vary the number of seasons of available per-cultivar data from one to 15 during training. Our results in Figure~\ref{fig:data_lim} show that the average error decreases across all model types, as more data becomes available. While biophysical models seem to perform better in the low data regime, the deep models perform better with more data. In contrast, DMC-MTL outperforms both methods across the entire range of data sizes.

\begin{figure}[t]
    \centering
    \includegraphics[width=\linewidth]{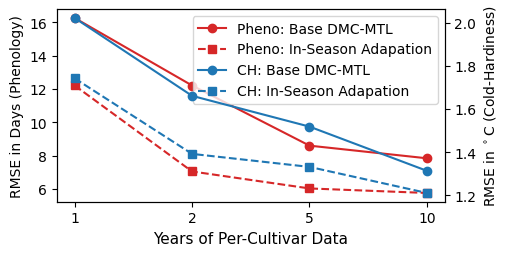}
    \caption{Performance of the base DMC-MTL with increasing per-cultivar data compared to the performance of the DMC-MTL with the additional in-season adaptation training for grape phenology (Pheno) and cold hardiness (CH). Results averaged over five seeds.}
    \label{fig:adaptation}
\end{figure}

\paragraph{Q3: In-Season Adaptation with In-Field Observations}
We evaluate the in-season adaptation approach against DMC-MTL by incorporating real-time in-field data. Using the same training data and protocol as the DMC-MTL models, we trained the in-season adaptation models with increasing amounts of per-cultivar data. Figure~\ref{fig:adaptation} shows the performance on the same testing data. By using the in-season observations to refine parameters predicted by the DMC-MTL model, the EEN yields the most benefit where DMC-MTL prediction errors are highest (the most data-limited settings). With only one season of data per cultivar, in-season adaptation reduces error by over 25\%, enabling more actionable forecasts. As DMC-MTL accuracy improves when additional data is available during training, the benefit of the EEN diminishes, indicating that in-season adaptation is most valuable when historical per-cultivar data is scarce. Furthermore, this approach is only viable when we cannot overfit the model to the training data as is the case when observations are subjective and noisy in the phenology and cold hardiness domains.

\begin{table*}[t]
\centering
\setlength{\tabcolsep}{1.5mm}
{\fontsize{9}{09}\selectfont
\begin{tabular}{lrrrr@{\hspace{15pt}}rrrr}
\toprule
            & \multicolumn{4}{c}{Phenology}  & \multicolumn{4}{c}{cold hardiness} \\
Approach  & \multicolumn{1}{c}{WA (Train Loc.)} & \multicolumn{1}{c}{VT} & \multicolumn{1}{c}{CA} & \multicolumn{1}{c}{OR} & \multicolumn{1}{c}{WA (Train Loc.)} & \multicolumn{1}{c}{VT} & \multicolumn{1}{c}{CA} & \multicolumn{1}{c}{OR}  \\
\midrule
DMC-MTL             & 5.9 $\pm$ 2.7 &  8.8 $\pm$  5.6 &  30.4 $\pm$ 9.3 & 17.7 $\pm$ 0.5 & 0.42 $\pm$ 0.28 & 0.76 $\pm$ 0.31 & 1.37 $\pm$ 0.16 & 3.59 $\pm$ 0.24\\
Deep-MTL                 & 6.1 $\pm$ 3.0 & 96.2 $\pm$ 10.8 & 120. $\pm$ 1.4  & 78.0 $\pm$ 1.9 & 0.34 $\pm$ 0.22 & 5.98 $\pm$ 1.57 & 6.01 $\pm$ 0.07 & 3.73 $\pm$ 0.46\\
PINN                & 5.3 $\pm$ 2.9 & 60.5 $\pm$ 14.2 &  59.8 $\pm$ 1.3 & 58.4 $\pm$ 3.9 & 0.39 $\pm$ 0.23 & 4.86 $\pm$ 1.41 & 8.38 $\pm$ 0.25 & 4.02 $\pm$ 0.40\\
TempHybrid          & 6.4 $\pm$ 5.2 & 97.3 $\pm$ 16.2 & 118. $\pm$ 20.  & 83.0 $\pm$ 18.& 4.25 $\pm$ 0.98 & 5.60 $\pm$ 0.98 & 8.57 $\pm$ 1.14 & 6.43 $\pm$ 1.33\\

\bottomrule
\end{tabular}
\caption{RMSE (days and $^\circ C$) for grape phenology and cold hardiness respectively, evaluated on unseen data sampled from the training location (WA) and from locations with a moderately similar weather (Vermont, California, and Oregon). Results averaged over five seeds.}
\label{tab:weather_results}
}
\end{table*}

\paragraph{Q4: Robustness to Differing Weather Conditions}
Current crop models are calibrated on a site specific basis, limiting their applicability to regions with sufficient historical data. Further, they assume that weather conditions will remain consistent and do not account for extreme weather events which is critical for broader adoption. 
To evaluate our approach's robustness to varying weather conditions, we trained models on synthetic data from Washington, USA, and evaluated them on data from Vermont, Oregon, and California, which have moderately similar weather patterns. Experiments were performed using synthetic phenology and cold hardiness datasets. Table~\ref{tab:weather_results} shows the cumulative RMSE in days on the Washington test set and Vermont, Oregon, and California test sets. While all models performed similar on the Washington test set, deep learning and other hybrid models produced large errors on the Vermont, Oregon, and California test sets. In contrast, DMC-MTL had a marginal increase in error on the test set, demonstrating robustness to varying weather conditions and its ability to produce usable predictions in unseen weather conditions. 

\begin{figure}[t]
    \centering
    \includegraphics[width=\linewidth]{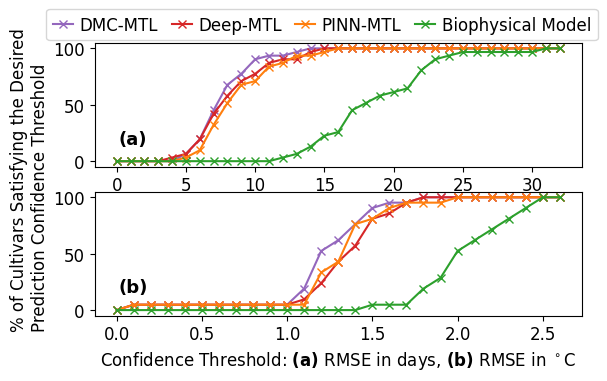}
    \caption{Percentage of all cultivars with cumulative error below a given RMSE threshold modeling a grape grower's tolerance for model prediction error. Results are reported over five seeds for (a) phenology and (b) cold hardiness.}
    \label{fig:rmse_alpha}
\end{figure}

\begin{figure}[t]
\centering
\includegraphics[width=0.95\linewidth]{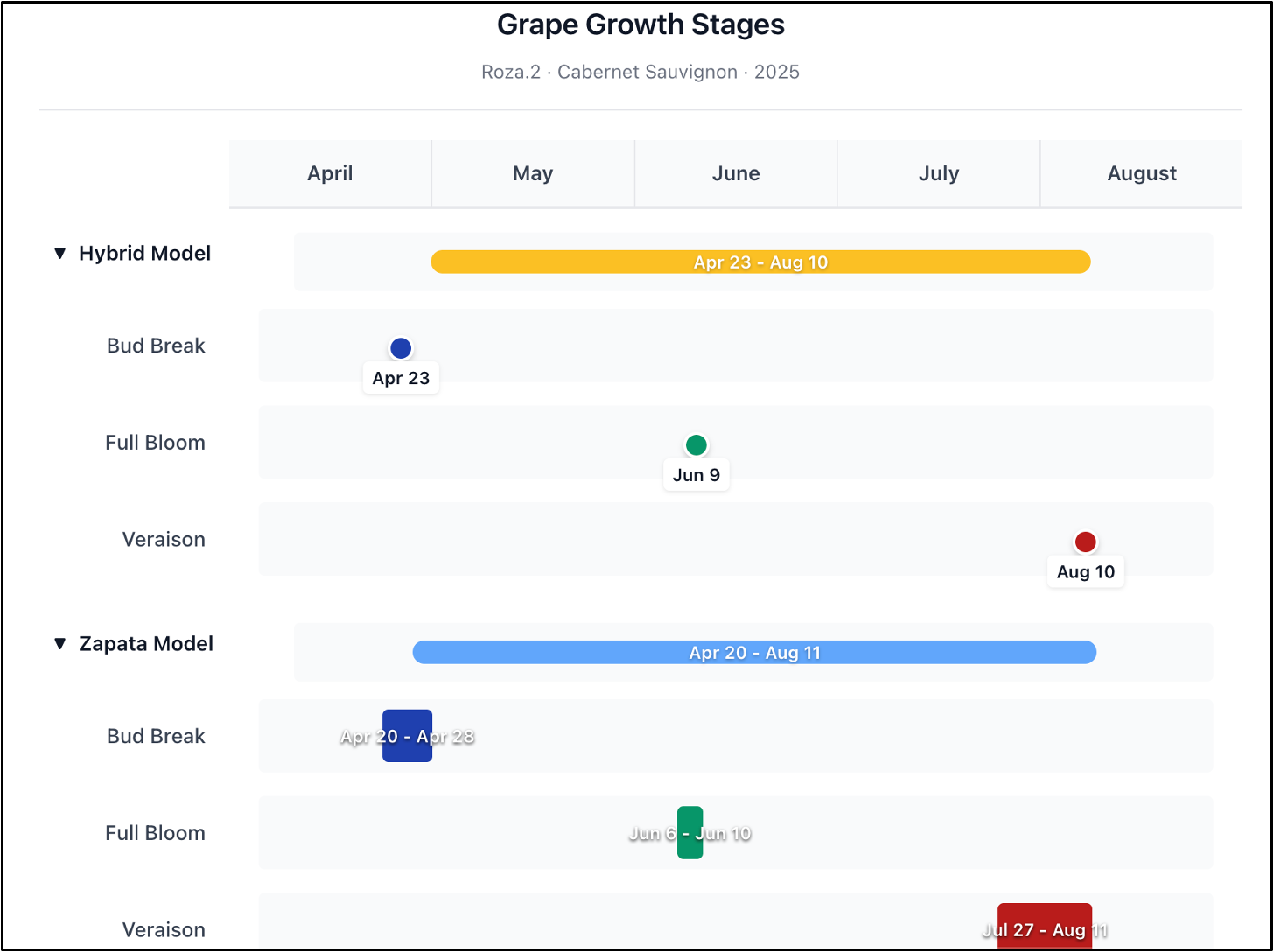}
\caption{Screenshot of the publicly accessible user interface on AgWeatherNet. The weather station-specific phenology forecasts during the 2025 season for the Cabernet Sauvignon cultivar of the Zapata (GDD-based biophysical model) are compared with the hybrid (DMC-MTL) model.  }
\label{fig:deployment}
\end{figure}

\paragraph{Q5: Accuracy of Per-Cultivar Predictions}
While DMC-MTL substantially reduces prediction error compared to biophysical models, its practical adoption depends on growers' error tolerance. To assess the potential model use, we evaluate the proportion of cultivars with RMSE below predefined thresholds, reflecting varying tolerance levels. We vary the RMSE tolerance threshold in the range $[0,2.5]$ for cold hardiness and $[0,30]$ for phenology. Our results in Figure~\ref{fig:rmse_alpha} show that DMC-MTL consistenly performs better than the baselines for both phenology and cold hardiness.  In addition, DMC-MTL's biologically realistic predictions position its medium range forecasts to be predicted unambiguously, positioning it to be widely used in the field.

\section{Deployment of DMC-MTL Models}
DMC-MTL models for grape phenology were recently deployed on AgWeatherNet~\cite{c:agnet} for the 2026 growing season. Figure~\ref{fig:deployment} is a screenshot of the user interface visible to growers, showing our model predictions for a specific weather station and grape cultivar (Cabernet Sauvignon) and that of the Zapata model (a GDD-based model that we compared against in Table ~\ref{tab:main_results})~\cite{zapata2017}. Stakeholders in the PNW widely use AgWeatherNet and will have access to both DMC-MTL and biophysical model forecasts to inform their vineyard operations which we will monitor for long-term impact. Continued analysis of deployed DMC-MTL models will provide insights to viticulturists on our interdisciplinary team to better understand the shortcomings of deployed biophysical models and improve understanding of grape phenological development. Outputs of all models are monitored for consistency and advice is posted in the case of rare climactic events that yield unexpected model predictions.

\section{Conclusion and Future Work}
We present a novel deep learning method that predicts the \textit{parameters} of biophysical models. Our results show that leveraging the benefits of both deep network architecture and biophysical models can outperform both methods individually.  In very data-limited settings, our in-season adaptation method provides growers with a more accurate prediction tool that is especially beneficial when waiting for more data is not feasible. Future work will aim to develop uncertainty quantification methods for crop state prediction with our hybrid modeling framework and expand predictions to additional regions.

\section*{Acknowledgments}
The authors thank Lynn Mills and Zilia Khaliullina at the Irrigated Agriculture Research and Extension Center (IAREC), Washington State University, for their invaluable support in collecting and sharing grape phenology data. The authors would also like to thank Sanjita Bhavirisetty, Jaitun Patel, and Dheeraj Vurukuti for their assistance in the deployment of DMC-MTL phenology models on AgWeatherNet. This research was supported by USDA NIFA award No. 2021-67021-35344 (AgAID AI Institute). 

\bibliographystyle{named}
\bibliography{ijcai26}

\appendix 
\section*{Appendix A: Model Architectures}
In this section we provide additional information on the DMC-MTL model and the variants that we used as experiment baselines. 
\subsubsection*{Appendix A.1: DMC Models}
The DMC-MTL model is composed of three parts, the RNN-Backbone (Figure~\ref{fig:model_arch}a), the multi-task embedding (Figure~\ref{fig:model_arch}b) and the interaction between the deep learning model and the biophysical model (Figure~\ref{fig:model_arch}c), encapsulating our proposed DMC-MTL approach. For all DMC architectures (DMC-MTL, DMC-STL, DMC-Agg, and multi-task variants), we used a Gated Recurrent Unit (GRU) with 1024 hidden units~\cite{chung2014}. The linear layers were reduced by factors of two: the first linear layer before the GRU was 256 hidden units, the second was 512 hidden units. After the GRU, the first linear layer was 512 hidden units and the second was 256 hidden units before making a prediction of the biophysical model parameters. Across all experiments, the embedding layer in $\mathcal{F}_\theta$ was the same size as the number of input features (16 for the real-world datasets and 11 for the synthetic datasets).

In contrast to DMC-MTL, DMC-STL and DMC-Agg did not have a multi-task embedding, and only utilized the RNN-backbone $f_\theta$ and the additional linear layer following $f_\theta$. DMC-STL models were trained using data from only a single cultivar. Meanwhile, DMC-Agg models were trained on unlabeled data aggregated across all cultivars. For our experiments, this meant that we trained five DMC-STL models per cultivar, whereas we only trained five DMC-Agg models per domain (i.e., phenology, cold-hardiness, wheat yield). 

\subsubsection*{Appendix A.2 Error Encoding Network Architecture}
The Error Encoding Networks (EENs) used in the in-season adaptation experiments use similar model architecture as DMC-MTL models. The input to the ENNs are the error signal (a scalar) and the cultivar i.d. $i$. In the same fasion as the DMC-MTL models, the one-hot embedding of the cultivar serves to distinguish between different error patterns that may vary between cultivars. However, given that the error signal is a scalar, the one-hot embedding is also a real-valued scalar, to avoid the cultivar embedding information dominating the error signal passed to $f_\theta$. The size of the first linear layer is adjusted to account for this change in input size, otherwise the layer sizes remain the same as DMC-MTL. 

If, during a growing season, the EEN observes no error between the observation and prediction made by the DMC-MTL model, then it should not modify the future parameters predicted by the DMC-MTL network. In this case, we can assume that the DMC-MTL model is predicting the observed data optimally, and no recalibration is needed. To accomplish this, the network layers of the EEN do not contain any bias terms, and when the observed error is zero, we zero out the one-hot cultivar embedding as well. As a result, the EEN learns to quickly recalibrate the parameters predicted by the DMC-MTL model (and thus adjust the internal state of the biophysical model) for future in-season predictions. 

~\citeauthor{henaff2017}~(\citeyear{henaff2017}) observed that in many cases, the desired prediction (in their case next video frame prediction) can be decomposed into a deterministic and stochastic portion, motivating the study of EENs to handle the stochastic prediction portion conditioned on previously observed error. In the crop state prediction task, the deterministic portion is explained by the weather and cultivar i.d. while we hypothesize that the stochastic portion can be explained by the sparsely observed error. Instead of learning an additive latent state, we instead learn parameters which are added to the predicted parameters of the pretrained DMC-MTL model that are then passed to the biophysical model to make the next crop state prediction. 

Given that the EEN and DMC-MTL model have different inputs, it is appropriate to train the EEN model on the same training data as the DMC-MTL model, while freezing the weights of the DMC-MTL model during EEN training. As a result, the EEN learns to correct the error made by a \emph{specific} DMC-MTL model. As shown in our results, this framing effectively reduces average prediction error in low data settings. To avoid relying on future in-season observations, during we randomly select a day $t$ for each season in the batch after which the EEN receives no further observations. 

\subsubsection*{Appendix A.3 Multi-Task Learning Variants}
In our experiments, we tested four approaches to embedding embedding task-specific cultivar information into our multi-task models. We expand on these methods below: 
\begin{enumerate}
    \item \textit{DMC-MTL} is our best performing model across all datasets and uses a concatenation embedding based on a one-hot embedding of the cultivar. This embedding is the dimensionality of the weather features (16 for real-world data and 11 for synthetic data) and is concatenated on to the weather data as shown in Figure~\ref{fig:model_arch}b before being fed into the first fully connected layer and subsequently the GRU.
    \item  \textit{DMC-MTL-Add} uses the same learnable one-hot embedding as DMC-MTL. However, this embedding vector is added to the weather vector instead of concatenated before being fed into the first fully connected layer. Thus, between DMC-MTL-Add and DMC-MTL, the first fully connected layer has a differing input size.
    \item  \textit{DMC-MTL-Mult} uses the same approach as DMC-MTL-Add, but instead multiplies the weather information by the learnable embedding vector. 
    \item  \textit{DMC-MTL-MultiH} uses different fully connected layers for each cultivar as prediction heads. This architecture shares the same structure as DMC-MTL (Figure~\ref{fig:model_arch}b) with the exception that there are 31 fully connected layers for phenology and 20 for cold-hardiness instead of a single output layer. Each final fully connected layer is trained only on the per-cultivar information, while the rest of the layers are trained in unison on all data. 
\end{enumerate}

These four approaches encompass hard parameter sharing and embedding methods fundamental to multi-task learning~\cite{zhang2018a}. Thus, along with DMC-STL and DMC-Agg variants described in the main text, they represent a comprehensive study on the utilty of multi-task learning in our proposed hybrid modeling framework.

\subsubsection*{Appendix A.4 Deep Learning Models}
The deep learning models (RNN with classification and regression targets) used the same network architecture as DMC-MTL ($\mathcal{F}_\theta$) as suggested by~\citeauthor{saxena2023a}~(\citeyear{saxena2023a}). However, instead of using 1024 hidden units for the GRU, we used 2048 hidden units and scaled the linear layers accordingly (see Figure \ref{fig:model_arch}b). The prediction target for the cold-hardiness and WOFOST wheat yield experiments (see \textit{Appendix B.4}) was a regression approximation of the crop state. For grape cold-hardiness prediction target was not just the LTE50, but also the LTE10 and LTE90 as well (the lethal temperatures at which 10\% and 90\% of dormant grape bud die, respectively) with the training loss as the sum of the MSE values across the LTE50, LTE10, and LTE90 observations. In practice, we found that this extra LTE10 and LTE90 data was unneeded to make accurate predictions of LTE50. For grape phenology we used a classification target of four classes corresponding to dormancy, bud break, bloom, and veraison. For the regression tasks we used mean squared error (MSE) loss and for the classification task we used Cross Entropy Loss with softmax activation to obtain the highest probability of a specific phenological stage. 

\subsubsection*{Appendix A.5 Hybrid Models}
\label{sec:PINN_arch}
\paragraph{Physics Informed Neural Network}
The PINN models used a network architecture identical to the deep learning models (Figure \ref{fig:model_arch}b) using a stage-based classification target for grape phenology and regression target for cold-hardiness and wheat yield. Unlike the deep learning models, the PINN models were trained using a physics-informed loss based on the biophysical model outputs to penalize biologically unrealistic predictions~\cite{aawar2025}:
\[L_{PINN}=\frac{1-p}{N}\sum_{i=1}^N(\hat{y_i}-y_i)^2+\frac{p}{N}\sum_{i=1}^N(\hat{y_i}-\dot{y_i})^2\]
where $y_i$ was the observed crop state, $\hat{y_i}$ was the crop state prediction of the PINN, and $\dot{y_i}$ was the crop state prediction of the biophysical model based on the best stationary model parameters. We found empirically that $p=0.5$ produced the best phenology predictions, and used that value in our cold-hardiness and wheat yield results. 

\paragraph{Residual Error Hybrid Model}
Residual error hybrid models combine a primary biophysical model with an auxiliary deep learning model to learn the structure of the remaining forecast error. Instead of assuming that the base biophysical model captures all predictive information, this hybrid approach explicitly learns the structure of the residual error with a deep learning approach. By reducing the scope that the data-driven model needs to learn, predictive accuracy and generalization is improved as the biophysical model provides the majority of the prediction structure.

We use the model architecture described in Figure~\ref{fig:model_arch}b. The biophysical model has per-cultivar parameters calibrated with Bayesian Optimization or brute force~\cite{solow2025,ferguson2014}. The output of the biophysical model is added to the output of the deep learning model and the sum of both terms is used in the loss function for training $\mathcal{F}_\theta$. 

\paragraph{TempResponse Hybrid Model}
The temperature response hybrid model for phenology~\cite{bree2025} replaces the temperature response function in the GDD phenology model with a small feedforward neural network. This design choice is based on the observation that many GDD model variants change the temperature response function which adds bias to the model. Instead~\citeauthor{bree2025}~(\citeyear{bree2025}) suggest to make this temperature response function learnable, reducing the bias in model choice selection. See \textit{Appendix C.1} for a description of the temperature response function. 

In the TempResponse hybrid model, the GDD model is implemented in a differentiable framework~\cite{paszke2017} and gradient descent is performed both on the neural network parameters and GDD model parameters to learn the best fit model parameters and temperature response function. 

While the TempResponse hybrid model was originally defined for tree phenology, we modified it to function with the Ferguson cold-hardiness model as well~\cite{ferguson2014}. The biophysical Ferguson cold-hardiness model models two stages of dormancy with a transition occurring when sufficient chilling units are reached~\cite{ferguson2011}. The chilling accumulation function is approximated by a small feedforward neural network in the same way that the temperature response function in the GDD model is learned. Thus, in the TempResponse hybrid model for cold-hardiness, we perform gradient descent on both the neural network parameters and Ferguson cold-hardiness model parameters to learn the best fit model. 

\section*{Appendix B: Biophysical Model Descriptions}
\label{sec:append_model_desc}

Before a biophysical model is used for agricultural crop state prediction, it must be calibrated with historical data. Common approaches used in the agricultural community for parameter calibration include brute force search~\cite{ferguson2014}, regression techniques~\cite{zapata2017}, and Bayesian optimization~\cite{seidel2018}. However, these approaches assume that a \emph{stationary parameter set} best explains the observed time series data during the growing season. Recent work has shown that early spring warming impacts phenology~\cite{guralnick2024a}, indicating that a stationary heat accumulation model is insufficient and it is well understood that other weather features impact phenology~\cite{greer2006,badeck2004}. Our approach addresses both of these concerns by conditioning phenology on exogenous weather features via daily parameter calibration. Below we describe the PyTorch implementation to convert biophysical models into a differentiable framework, the biophysical models, and the parameters of the models that are calibrated using our DMC-MTL approach.

\subsection*{Appendix B.1: PyTorch Implementation}
To create differentiable implementations, we replace all mathematical operation in each biophysical model with the corresponding PyTorch operation so that gradients are tracked. Parameters, states, and rates are instantiated as tensors instead of floats. To enable batch learning, all conditional statements are replaced by `where' statements.

\begin{table*}[ht!]
\centering
\setlength{\tabcolsep}{1mm}
{\fontsize{9}{09}\selectfont
\begin{tabular}{lllllrr}
\toprule
                                 & Parameter Name & \multicolumn{2}{l}{Parameter Description}                          & Unit                   & \multicolumn{1}{l}{Min Value} & \multicolumn{1}{l}{Max Value} \\
                                 &                &                                                      &             &                        & \multicolumn{1}{l}{}          & \multicolumn{1}{l}{}          \\
                                 \midrule
\multirow{7}{*}{GDD Model}       & TBASEM         & \multicolumn{2}{l}{Base Temperature ($T_b$)}                       & $^\circ C$             & 0                             & 15                            \\
                                 & TEFFMX         & \multicolumn{2}{l}{Maximum Effective Temperature ($T_m$)}          & $^\circ C$             & 15                            & 45                            \\
                                 & TSUMEM         & Temperature Sum for Bud Break                        &             & $^\circ C$             & 10                            & 100                           \\
                                 & TSUM1          & Temperature Sum for Bud Break                        &             & $^\circ C$             & 100                           & 1000                          \\
                                 & TSUM2          & Temperature Sum for Bloom                            &             & $^\circ C$             & 100                           & 1000                          \\
                                 & TSUM3          & Temperature Sum for Veraison                         &             & $^\circ C$             & 100                           & 1000                          \\
                                 & TSUM4          & Temperature Sum for Ripening                         &             & $^\circ C$             & 100                           & 1000                          \\
                                 \midrule
\multirow{10}{*}{Ferguson Model} & HCINIT         & Initial Cold-Hardiness                               &             & $^\circ C$             & -15                           & 5                             \\
                                 & HCMIN          & Minimum Cold-Hardiness                               &             & $^\circ C$             & -5                            & 0                             \\
                                 & HCMAX          & Maximum Cold-Hardiness                               &             & $^\circ C$             & -40                           & -20                           \\
                                 & TENDO          & Base Temperature During Endodormancy                 &             & $^\circ  C$            & 0                             & 10                            \\
                                 & TECO           & Base Temperature During Ecodormancy                  &             & $^\circ C$             & 0                             & 10                            \\
                                 & ENACCLIM       & Acclimation Rate During Endodormancy                 &             & $^\circ C^\circ C^{-1}$ & 0.2                           & 0.2                           \\
                                 & ECACCLIM       & Acclimation Rate During Ecodormancy                  &             & $^\circ C^\circ C^{-1}$ & 0.2                           & 0.2                           \\
                                 & ENDEACCLIM     & Deacclimation Rate During Endodormancy               &             & $^\circ C^\circ C^{-1}$ & 0.2                           & 0.2                           \\
                                 & ECDEACCLIM     & Deacclimation Rate During Ecodormancy                &             & $^\circ C^\circ C^{-1}$ & 0.2                           & 0.2                           \\
                                 & ECOBOUND       & Threshold for Ecodormancy Transition                 &             & $^\circ C$             & -800                          & -200                          \\
                                 \midrule
\multirow{7}{*}{WOFOST Model}    & DLO            & \multicolumn{2}{l}{Optimum Daylength for Development}              & Hours                  & 12                            & 18                            \\
                                 & TSUM1          & \multicolumn{2}{l}{Temperature Sum for Anthesis}                   & $^\circ  C$            & 500                           & 1500                          \\
                                 & TSUM2          & \multicolumn{2}{l}{Temperature Sum for Maturity}                   & $^\circ  C$            & 500                           & 1500                          \\
                                 & VERNBASE       & \multicolumn{2}{l}{Base Vernalization Requirement}                 & Days                   & 0                             & 25                            \\
                                 & VERNSAT        & \multicolumn{2}{l}{Saturated Vernalization Requirement}            & Days                   & 0                             & 100                           \\
                                 & CVO            & \multicolumn{2}{l}{Storage Organ Conversion Efficiency}            & $kg\cdot kg^{-1}$      & 0.5                           & 0.8                           \\
                                 & RMO            & \multicolumn{2}{l}{Storage Organ Relative Maintenance Respiration} & ---                   & 0.05                          & 0.2                           \\
                                 \bottomrule
\end{tabular}
}
\caption{The parameters of the GDD Model, Ferguson Model, and WOFOST model used in the DMC-MTL approach. The ranges correspond to the minimum and maximum values that the parameter can be after $\tanh$ activation normalizing from the range $[-1,1]$}
\label{tab:model_params}
\end{table*}

\subsection*{Appendix B.2: Growing Degree Day Phenology Model}
Grape phenology is described by the Eichhorn-Lorenz phenological stages~\cite{lorenz1995} and includes three key phenological states: bud break, bloom, and veraison. Accurate prediction of these three states enable growers to follow crop management policies more precisely in order to increase yield and quality, and to increase vineyard efficiency by ensuring farm labor is available during the growing season. 

The Growing Degree Day (GDD) model is a mechanistic phenology model that makes predictions from January 1st until September 7th, for the three key phenological stages~\cite{zapata2017}. The GDD model accumulates the amount of Degree Days (DD) needed to transition between phenological stages. Given a base temperature value $T_b$, and a maximum effective temperature $T_m$, the degree days can be computed:

\[DD=\sum_{i=1}^H\min\big(T_m, (T_i-T_b)\big)\]

where $H$ is the length of the season and the term inside the summand is the temperature response function. A stage transition occurs when DD is greater than a specific threshold. Each stage, bud break, bloom, and veraison, has an associated threshold value. In Table \ref{tab:model_params} we list the seven parameters of the GDD model and the associated ranges that we chose to use in the DMC method for normalizing parameters after the $\tanh$ activation. 

\subsection*{Appendix B.3: Ferguson Cold-Hardiness Model Description}
Grape cold-hardiness characterizes the grapevine's resistance to lethal cold temperatures from September 7th to May 15th~\cite{ferguson2011}. When cold-hardiness is low in the spring and fall, sudden frost events can cause significant damage to to dormant buds resulting in a decrease in yield quantity. Cold-hardiness is difficult to measure in the field; consequently, grape growers rely on the Ferguson model for daily predictions of LTE50, the temperature when 50\% of dormant buds freeze~\cite{ferguson2014}. By contrasting the LTE50 predictions with the weather forecast, grape growers decide whether preventative measures (e.g., wind machines and heaters) are needed to protect the dormant buds. The Ferguson model computes the change in LTE50, $\Delta H_c$, as a function of daily acclimation and deacclimation based on dormancy stage and ambient temperature. See~\citeauthor{ferguson2011}~(\citeyear{ferguson2011}) for a complete description. The Ferguson model parameters that we calibrate in our approach and the corresponding ranges are listed in Table \ref{tab:model_params}.

\subsection*{Appendix B.4: WOFOST Wheat Model Description}
As an additional testing domain to complement phenology and cold-hardiness, we consider the wheat yield domain. Accurate forecast of staple crop yields are critical to financial planning.
The WOFOST crop growth model~\cite{vandiepen1989} is widely used to predict field level yield for many crops, including winter wheat, by predicting the daily yield (as the daily weight of the storage organs) from January 1st to September 1st each year. 

Predicting hectare-level wheat yield is critical for economical planning~\cite{allen1994}. Using historical weather data, the WOFOST model can generate synthetic wheat yield observations. The WOFOST model is a significantly more complicated model compared to the phenology and cold-hardiness models. While creating a PyTorch implementation required upfront work, our results demonstrate that our hybrid approach significantly outperforms other modeling approaches for the prediction of winter wheat yield during the growing season. 

\begin{figure*}[t]
    \centering
    \includegraphics[width=\linewidth]{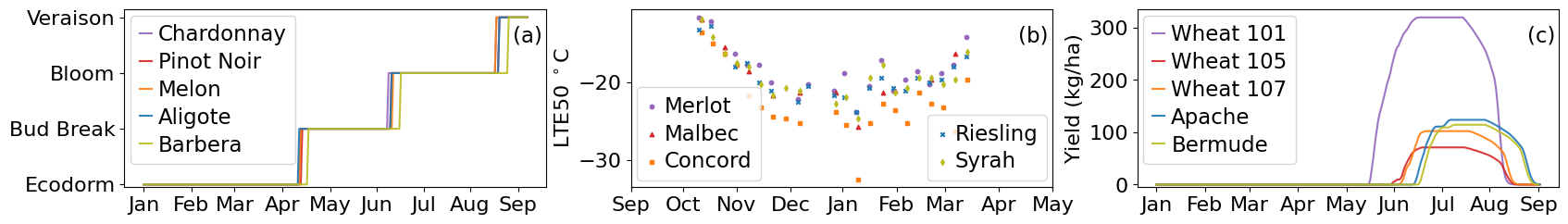}
    \caption{Daily crop state observations for five cultivars of (a) grape phenology and (b) grape cold-hardiness during a single growing season. Modeling approaches must predict these curves with biologically consistency. Despite experiencing the same weather, cultivars exhibit different behaviors, making naive data aggregation inadequate and motivating the use of a multi-task approach.} 
    \label{fig:char_of_data}
\end{figure*}

\section*{Appendix C: Datasets Details}
In this section we discuss the real-world data used in our experiments. 
\subsection*{Appendix C.1: Real World Data}
\label{sec:real_world_data}
As mentioned in the main text, our experiments are conducted in a data-limited setting. The cold-hardiness and phenology from up to 32 genetically diverse cultivars/genotypes of field-grown grapevines has been measured since 1988 in the laboratory of the WSU Irrigated Agriculture Research and Extension Center in Prosser, WA (46.29°N latitute; -119.74°W longitude). In the vineyards of the IAREC, the WSU-Roza Research Farm, Prosser, WA (46.25°N latitude; -119.73°W longitude), and in the cultivar collection of Ste. Michelle Wine Estates, Paterson, WA (45.96°N latitute; -119.61°W longitude), cane samples containing dormant buds were collected daily, weekly, or at 2-week intervals from leaf fall in autumn to bud swell in spring while phenological stage observations (onset of bud break, bloom, and veraison) were observed during from bud swell to leaf fall. Phenology was observed daily. 

Table \ref{tab:real_pheno_data} shows a summary of the number of years of phenology data and the number of cold-hardiness samples collected per cultivar after data processing. In addition to the cold-hardiness and phenology measurements, the real-world weather data from nearby open-field weather station was used and  contains 14 weather features: date; min, max and average temperature, humidity, and dew point; solar irradiation; rainfall; wind speed; and evapotranspiration. The synthetic datasets generated from the NASAPower database contain nine weather features: date, day length, min, max, and average temperature, reference and potential evapotranspiration, rainfall, and solar irradiation. The processed datasets are available in our code repository \url{https://tinyurl.com/IRAS-DMC-MTL}, and the raw data may be shared upon request. 

\begin{table}[t]
\centering
\setlength{\tabcolsep}{1mm}
{\fontsize{9}{09}\selectfont
    \begin{tabular}{lrrrr}
    \toprule Cultivar & \makecell{Years of\\Pheno. Data} & \makecell{Years of\\LTE Data} & \makecell{Total LTE\\Samples} \\
    \midrule
Aligote               &  9 &  2 &  20 \\ 
Alvarinho             &  9 & 10 & 120 \\ 
Auxerrois             &  9 &  8 & 101 \\ 
Barbera               &  8 & 11 & 130 \\ 
Cabernet Franc        & 17 &  3 &  28 \\ 
Cabernet Sauvignon    & 18 & 27 & 629 \\ 
Chardonnay            & 21 & 20 & 593 \\ 
Chenin Blanc          & 17 & 15 & 160 \\ 
Concord               & 16 & 20 & 403 \\ 
Durif                 &  9 &  0 &   0 \\ 
Gewurztraminer        & 16 &  7 &  78 \\ 
Green Veltliner       &  9 & 10 & 120 \\ 
Grenache              & 15 & 13 & 144 \\ 
Lemberger             & 17 &  4 &  43 \\ 
Malbec                & 17 & 14 & 208 \\ 
Melon                 &  8 &  1 &  10 \\ 
Merlot                & 21 & 20 & 797 \\ 
Mourvedre             &  9 & 10 & 118 \\ 
Muscat Blanc          & 16 & 10 & 119 \\ 
Nebbiolo              &  8 & 13 & 153 \\ 
Petit Verdot          &  8 & 10 & 117 \\ 
Pinot Blanc           &  9 &  6 &  74 \\ 
Pinot Gris            & 18 & 13 & 148 \\ 
Pinot Noir            & 17 & 10 & 121 \\ 
Riesling              & 17 & 27 & 524 \\ 
Sangiovese            &  9 & 13 & 148 \\ 
Sauvignon Blanc       &  9 & 12 & 141 \\ 
Semillon              & 17 & 12 & 186 \\ 
Syrah                 &  2 & 17 & 414 \\ 
Tempranillo           &  8 &  7 &  81 \\ 
Viognier              &  9 & 12 & 147 \\ 
Zinfandel             & 13 & 12 & 133 \\ 
    \bottomrule
    \end{tabular}
}    
\caption{Summary of real-world grapevine cultivar phenology and cold-hardiness observations collected from Washington State University in Prosser, WA. }
\label{tab:real_pheno_data}
\end{table}

\subsection*{Appendix C.2: Data Processing} 
Historical grapevine data is inherently noisy and contains many missing weather observations. To make the data usable, we process it in the following ways: (1) If any weather feature is missing more than 10\% values, we discard the entire season. Otherwise, we fill missing values with linear interpolation between the two nearest observed values. (2) We normalize all weather features using z-score normalization. For the date, we use a two feature periodic date embedding using sine and cosine. (3) For phenology, we discard any seasons that do not record bud break, bloom, and veraison. We fill values between observations with the last previous observation, as only the onset of a phenological stage is recorded in the dataset. We ignore other phenological stages present as they are not predicted by the GDD model. (4) For cold-hardiness, we include any season with at least one valid LTE50 observation. Missing LTE50 values are masked during training.

For our phenology experiments, we consider all cultivars except Syrah as there is not sufficient data to form a test set. For our cold-hardiness experiments, we omit the Aligote, Alvarinho, Auxerrois, Cabernet Franc, Durif, Green Veltliner, Melon, Muscant Blanc, Petit Verdot, Pinot Blanc, Pinot Noir, and Tempranillo cultivars from our dataset either due to insufficient data for a test set, or inavailability of Ferguson model parameters to serve as a baseline. 

\subsection*{Appendix C.3: Characteristics of Crop Data}
Real-world crop observation data are governed by strict biological constraints. For example, phenological observations resemble a step function and cannot return to a previous stage. Wheat yield observations are a strictly concave curve: yield increases during the reproductive phase and decreases after the crop ripens until death. Figure~\ref{fig:char_of_data} illustrates the structured nature of seasonal observations in grape phenology, cold-hardiness, and wheat yield data across five cultivars. Prediction approaches that violate these constraints and produce biologically unrealistic outputs, including those with low average error, cannot be trusted for medium-range forecasting~\cite{rudin2019}. Furthermore, seasonal observations are sparse and often vary per cultivar, requiring efficient data aggregation for learning. 

All three domains in Figure~\ref{fig:char_of_data} share the following characteristics: data is sparse among cultivars, values have a strict biological structure, and observations are infrequent or unchanging for a large portion of the growing season. These shared characteristics make the cold-hardiness and wheat yield domains valuable benchmarks for evaluating our proposed hybrid modeling framework. While conventional classification and regression approaches may seem appropriate, our results show they frequently produce biologically inconsistent outputs and higher prediction errors. In contrast, our proposed dynamic parameter calibration approach achieves lower average error while maintaining biological consistency, offering a more reliable solution for real-world crop state forecasting to inform agricultural decision making.

\begin{table}[t]
\centering
\setlength{\tabcolsep}{1mm}

\begin{tabular}{lrrrr}
\toprule
           \makecell{Models}      & \makecell{Hidden\\Size} & \makecell{Learning\\Rate} & \makecell{Batch\\Size} & \makecell{Learning Rate\\Anneal} \\
\midrule
DMC-MTL          & 1024                            & 0.0001                            & 12                             & 0.9                           \\
DMC-STL          & 1024                            & 0.005                             & 4                              & 0.9                           \\
RNN              & 2048                            & 0.0001                            & 12                             & 0.9                           \\
PINN             & 2048                            & 0.0001                            & 12                             & 0.9                           \\
TempHybrid       & 64                              & 0.02                              & 4                              & 0.9                           \\
Gradient Descent & N/A         & 0.1                               & 4                              & 0.9                           \\
\bottomrule
\end{tabular}
\caption{Best hyperparameters found for each model type after five-fold cross validation on the entire grape phenology dataset.}
\label{tab:hyperparameters}
\end{table}

\begin{table}[t]
\centering
\setlength{\tabcolsep}{4mm}
{\fontsize{9}{09}\selectfont
\begin{tabular}{llr}
\toprule
&      Solution Approaches           & Synthetic Wheat Yield \\
\midrule
\midrule

\multirow{5}{*}{\textbf{Q1a:}} 
                     & \textbf{DMC-MTL (Ours)}       & \textbf{10.63 $\pm$ 7.39}\phantom{$^*$} \\
                     & Gradient Descent              & 12.69 $\pm$ 10.7$^*$ \\
                     & Deep-MTL                      & 31.63 $\pm$ 16.8$^*$  \\
                     & PINN                          & 36.56 $\pm$ 18.8$^*$  \\
                     & Residual Hybrid               &  14.11 $\pm$ 10.9$^*$\\ 
                     \midrule
\multirow{5}{*}{\textbf{Q1b:}} 
                     & DMC-STL                              & 15.46 $\pm$ 17.1$^*$  \\
                     & DMC-Agg                              & 42.29 $\pm$ 7.58$^*$  \\
                     & DMC-MTL-Mult                          & 14.42 $\pm$ 9.74$^*$  \\       
                     & DMC-MTL-Add                           & 13.66 $\pm$ 9.96$^*$  \\       
                     & DMC-MTL-MultiH                        & 14.72 $\pm$ 8.92$^*$  \\  
\bottomrule
\end{tabular}
}
\caption{The average seasonal error (RMSE in $kg/ha$ for wheat yield) over all ten cultivars and five seeds in the testing set. DMC-MTL is compared against gradient descent on the biophysical model, a deep learning approach, two hybrid models, and five DMC variants. Best-in-class results are reported in bold. A $^*$ indicates that DMC-MTL yields a \textit{statistically significant improvement} ($p < 0.05$) using the paired t-test relative to the corresponding baseline. }
\label{tab:main_results}
\end{table}

\section*{Appendix D: Experimental Details and Hyperparameter Selection}
We outline additional portions of our experimental protocol not included in the main text for reproducibility. All experiments were run on a Ubuntu 24.04 system with a NVIDIA 3080Ti with 10GB of VRAM. 
As noted in the main text and in \textit{Appendix C.1}, the amount of per-cultivar data is limited, there is high variance in weather and phenological response each season, and some cultivars only have four seasons of data. To address these challenges in validation and evaluation, we build our training and validation sets as follows. 

For five different seeds, we first withhold two seasons of data per cultivar for the testing set. Then, we withhold an additional season of data from the training per cultivar for the validation set. For the most data scarce cultivars, this results in one season of data for the training set and one season for the validation set. However, many cultivars have more training data that is leveraged given the multi-task setting. 

We consider five different hyperparameters for our 5-fold validation. (1) Number of GRU hidden units in $[128, 256, 512, 1024, 2048]$, (2) Learning rate in $[0.01, 0.005, 0.001, 0.0005, 0.0001]$, (3) Batch size in $[4, 12, 16, 24, 32]$, (4) Learning rate annealing in $[0.8, 0.85, 0.9, 0.95, 1]$, and (5) Coefficient $p$ for PINN loss in $[0.1, 0.25, 0.5, 0.75, .9]$. We also considered the omission of the extra linear layers before the GRU. 

We performed a grid search over these parameters for all model types. The average performance on the validation set for each set of hyperparameters was recorded over the five training/validation/testing splits. After these hyperparameters (see Table~\ref{tab:hyperparameters}) were selected, we retrained each model five times using the combined training and validation data from each split with the best set of hyperparameters found.

\section*{Appendix E: Additional Results}
We consider additional research questions to supplement the five research questions in the main text. Our questions are motivated by further understanding the impact of multi-task learning, the importance of expressivity and history in our hybrid modeling approach, and leveraging these observations to inform viticultural research. These questions are: 
\begin{enumerate}
\setlength{\itemsep}{0cm}
    \item[\q{1}:]{In the wheat yield setting, how does DMC-MTL compare to (a) baselines and (b) alternative multi-task embeddings?}
    \item[\q{2}:]{Do per-cultivar predictions improve under the multi-task learning framework?}
    \item[\q{3}:]{How well does DMC-MTL optimize per-stage phenology predictions?}
    \item[\q{4}:]{Does the amount that the daily parameter predictions are allowed to vary impact DMC-MTL? }
    \item[\q{5}:]{What is the importance of previous weather in daily parameter predictions?}
    \item[\q{6}:]{How can daily parameter predictions inform viticultural research?}
\end{enumerate}

\paragraph{Q1: WOFOST Wheat Yield}
Supporting our results in grape phenology and cold-hardiness, DMC-MTL achieved best in class results on the WOFOST wheat yield dataset. It out performed gradient descent on the seven model parameters (see Table~\ref{tab:model_params}), the multi-task deep learning model, and other hybrid modeling approaches. This result demonstrates DMC-MTL's robustness to different prediction tasks, given the complexity of the WOFOST model and even outperforms other the gradient descent baseline, despite the fact that the synthetic data was generated using static parameters. 

Additionally, the WOFOST multi-task results (Q1b in Table~\ref{tab:main_results}) mirror those reported in the main text for phenology and cold-hardiness. In the WOFOST results, the tasks vary more widely, as evidenced by the poor performance of DMC-Agg. Nonetheless, DMC-MTL is still an improvement over DMC-STL, indicating our approach's ability to share limited data efficiently across tasks. DMC-MTL also outperformed the other methods of providing task-specific information. 

\begin{table}[t]
\centering
{\fontsize{10}{09}\selectfont
\begin{tabular}{lrr}
\toprule
                & Phenology & Cold-Hardiness  \\
\midrule
\midrule
\% Improve & 84\%   & 100\% \\
Avg. Increase     & 2.41 $\pm$ 1.88            & 0.37 $\pm$ 0.15           \\
Avg. Decrease   & 0.51 $\pm$ 0.28            & 0.00 $\pm$ 0.00     \\
\bottomrule
\end{tabular}
}
\caption{Comparison of DMC-MTL vs DMC-STL on a per-cultivar basis with the percentage of cultivars that saw a reduction in error from DMC-STL to DMC-MTL along with the average reduction in error (RMSE in days for phenology, $^\circ C$ for cold-hardiness) and the average increase in error for cultivars where accuracy worsened.}
\label{tab:data_sharing}
\end{table}

\paragraph{Q2: Per-Cultivar Predictions}
Table~\ref{tab:data_sharing} shows the per-cultivar comparison between DMC-MTL and DMC-STL. For phenology, 84\% of cultivars on average saw a reduction of error using DMC-MTL compared to DMC-STL with an average decrease in error by 2.41 days while for the 16\% of cultivars that saw an increase in error, that increase was marginal at only 0.51 days. Meanwhile for cold-hardiness, all cultivars saw a decrease in error. Most interestingly, even cultivars with larger datasets still see an increase in performance compared to the DMC-STL models per-cultivar. This result indicates that even these larger cultivar datasets are insufficient to fully capture the underlying patterns within the data and thus benefit from data aggregation with additional cultivar datasets. Overall, the results demonstrate that DMC-MTL leverages data efficiently across cultivars and generally improves per-cultivar predictions, enabling accurate prediction with limited data. 

\begin{figure}
    \centering
    \includegraphics[width=\linewidth]{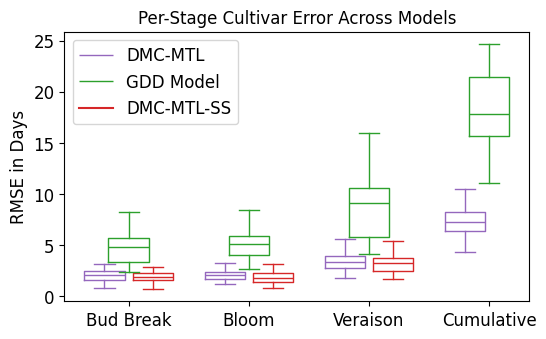}
    \caption{The distribution of per-stage prediction error (RMSE in days) of the DMC-MTL model and GDD model. Additionally, DMC-MTL-SS models were trained to minimize per-stage error (as opposed to cumulative stage error).}
    \label{fig:per_stage_error}
\end{figure}

\paragraph{Q3: Optimization of Per-Stage Phenology Predictions}
DMC-MTL demonstrated a reduction in cumulative error across the three key phenological stages (Table 2 in main text); however, for grape growers to effectively use the model, it is important to understand how well DMC-MTL minimized error at each individual stage. As a baseline (DMC-MTL-SS), we trained DMC-MTL models on the same-real world grape phenology dataset, but changed the objective to minimize only the prediction error of a single stage: bud break, bloom, or veraison. In Figure~\ref{fig:per_stage_error}, we show the average error across cultivars attributed to each stage. 

Our results show that DMC-MTL effectively minimized error in predicting bud break, bloom and veraison stages, performing similar to the single-stage prediction baseline DMC-MTL-SS. Both the DMC-MTL and GDD models exhibited similar trends in the difficulty of prediction; bud break and bloom had similar errors, while veraison proved to be harder to predict. However, these results were near identical to the DMC-MTL-SS baseline, indicating that the DMC-MTL model was able to optimize over all stages effectively without compromise. The variance can be attributed to different cultivars; we found that the data from some cultivars is inherently harder to predict accurately. 
\paragraph{Q4: Effect of Paramter Smoothing}
A key assumption of DMC-MTL is that the model parameters can vary in the ranges given by Table~\ref{tab:model_params}. This assumption ensures that predictions remain biologically realistic by staying within known parameter ranges, but also allows for increased expressivity when compared to the stationary parameter biophysical models. However, the stationary parameter biophysical models and DMC-MTL hybrid model exist on opposite sides of a spectrum of parameter smoothness. DMC-MTL allows for large changes in daily parameter predictions while the stationary biophysical model disallows any parameter change. 

To investigate the importance of the increased model expressiveness that DMC-MTL allows, we consider DMC-MTL variants that limit the daily change in parameters. To do so, we let the DMC-MTL GRU predict a daily delta parameter term and add this delta to the previous day's parameters before using the biophysical model to predict the crop state. To force predictions to remain biologically realistic, we clamp the ranges given in Table~\ref{tab:model_params}. We let the model parameters vary by a factor of 0.01, 0.001, 0.0001, and 0.00001 of the original parameter ranges. We choose 0.01 as the largest factor as it roughly represents the largest observed daily change in the base DMC-MTL models. 

\begin{figure}[t]
    \centering
    \includegraphics[width=\linewidth]{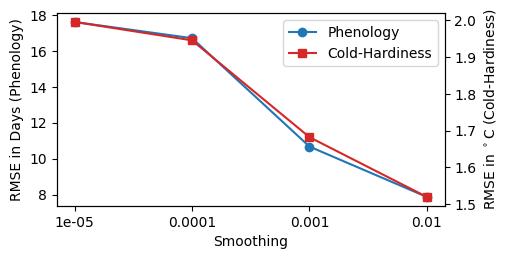}
    \caption{The performance of DMC-MTL with decreased parameter smoothing. Results are averaged over five seeds}
    \label{fig:smoothing}
\end{figure}

Our results in Figure~\ref{fig:smoothing} demonstrate that increased expressivity (model parameter delta scale increasing from 0.00001 to $0.01$) increase the performance of the DMC-MTL model. At a parameter smoothness of 0.00001, model performance roughly matches the GDD and Ferguson biophysical models. However, by a parameter smoothing factor of 0.01, performance approaches that of the base DMC-MTL models. Thus, we conclude that varying model parameters daily and in large ranges is critical for the accuracy of our modeling approach. 

\paragraph{Q5: Impact of Increasing Weather Windows on Parameter Selection}
The DMC-MTL approach leverages a RNN to encode the weather information for the growing season up to day $T$ to predict biophysical model parameters for day $T$. However, the biophysical also maintain an internal crop state that evolves with each input. Thus, unlike a deep learning model which will require the entire weather sequence to make adequate phenology predictions, our hybrid DMC-MTL model may not. To demonstrate the importance of the recurrent modeling choice in encoding this previous weather information in the DMC-MTL framework, we consider additional variants to the approach where the GRU uses a sliding window $k$ and only uses the weather featuers from the previous $T-k$ days to make parameter predictions for day $T$. We also compare against a non-recurrent network architecture that replaces the  with three feed forward layers, each with 64 hidden units, with ReLU activation. 

\begin{figure}[t]
    \centering
    \includegraphics[width=\linewidth]{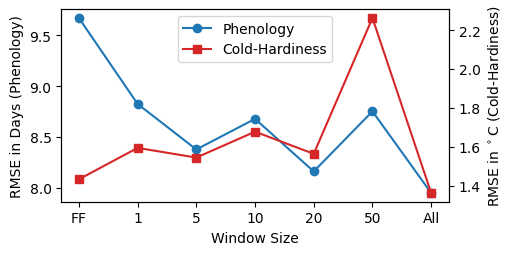}
    \caption{The performance of DMC-MTL with a sliding weather window. Results are averaged over five seeds. }
    \label{fig:weather_window}
\end{figure}

Our results in Figure~\ref{fig:weather_window} show that for phenology, there is a general trend where a larger weather window increases the prediction accuracy. This trend is not as clear in cold-hardiness. However in cold-hardiness prediction, many other models only use the previous three or five days~\cite{salazar-gutierrez2020}, indicating that previous weather is not as critical. Nonetheless, the best model performance still utilized all weather information. Overall, we can conclude that even though the DMC-MTL hybrid approach uses a biophysical model that maintains an internal state, there is still value in incorporating previously seen weather information into daily predictions.  

Between the results in Figures~\ref{fig:smoothing} and~\ref{fig:weather_window} it is clear that varying model parameters and a larger weather window increases the prediction accuracy of DMC-MTL compared to its variants as well as the biophysical GDD and Ferguson grape phenology and cold-hardiness models. These observations are in line with viticultural research that early season weather and non-stationary parameterization are key to proper phenology modeling~\cite{greer2006,guralnick2024a}. Without the use of deep learning techniques and a hybrid approach, it is not immediately clear how to accomplish these two desiderata for viticulturalists. Thus, DMC-MTL offers a compelling approach to accurate phenology modeling. 

\paragraph{Q6: Interpretability of DMC-MTL}
Interpretable models are desirable for agronomists as they increase trust and further scientific understanding of underlying crop processes~\cite{rudin2019}. Unlike purely data-driven deep learning models, our hybrid modeling approach enables sensitivity and attribution analysis through methods such as integrated gradients~\cite{sundararajan2017}. Observing the results of the integrated gradients analysis enables viticulturalists to understand how exogenous weather features impact the prediction of different parameters withing the biophysical model. As these model parameters and the biophyisical model are well understood, hypotheses can be generated on the impacts of these exogenous weather features on phenology which can then be verified through laboratory and field trials.

\begin{figure}[t]
    \centering
    \includegraphics[width=\linewidth]{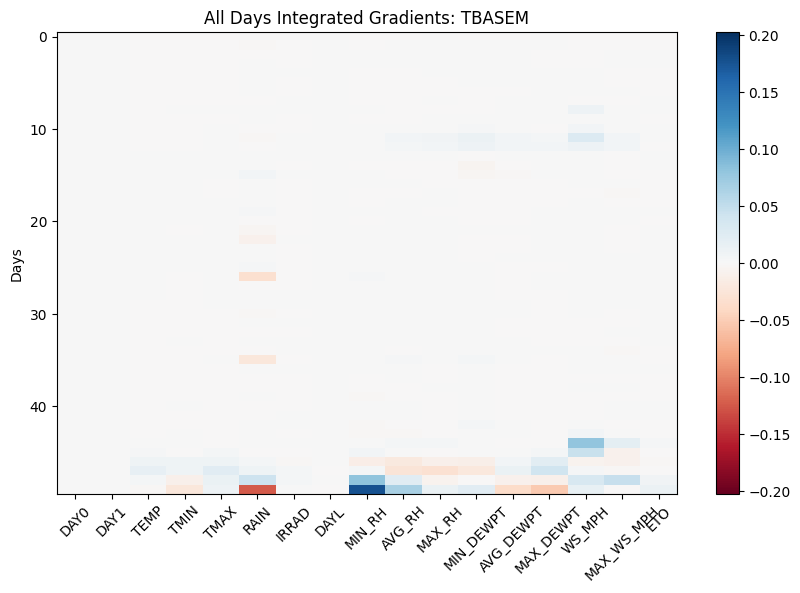}
    \caption{The integrated gradients values of the Base Temperature Sum parameter prediction for the 2024 growing season of the Chardonnay cultivar. Red values indicate positive attribution and blue values indicate negative attribution.}
    \label{fig:ig}
\end{figure}

In Figure~\ref{fig:ig}, we show the integrated gradients analysis of the 17 input features in the DMC-MTL model and their impact on the prediction of the Base Temperature Sum parameter for day 50 in the 2024 growing season for the Chardonnay cultivar for phenology prediction. First, observe that the majority of the attribution is clustered within the six days prior to the prediction. This is intuitive as more recent weather data should be more relevant. However, we also see attribution as far back as day seven for the dew point and wind features. This further validates our findings in Figure~\ref{fig:weather_window} that historical data is broadly important to accurate phenology predictions. 

Interestingly, rainfall and relative humidity have the largest negative and positive impacts on the prediction value of the Base Temperature Sum parameter relative to the baseline. In this experiment, the baseline is the daily average of each weather feature across all observed weather. However, other baselines such as the all zeroes baseline are possible choices, with the latter being more popular in attribution in image classification networks~\cite{sundararajan2017}. Taking these observations into account, a viticulturalist can then use the integrated gradients results of a trained DMC-MTL model to hypothesize and validate effects of exogenous weather features on phenology. 

This analysis is only made possible by the hybrid modeling approach. A data-driven approach does not give sufficient insight into the underlying crop process, while the available biophysical models do not account for the impacts of exogenous weather features. DMC-MTL is a best-of-both worlds in this space, offering increased prediction accuracy over the biophyisical model and increased interpretability compared to data-driven approaches. 

\paragraph{Summary of Results}
In this section, we have expanded on our grape phenology and cold-hardiness results with the addition of the WOFOST wheat yield domain. Our results demonstrate that DMC-MTL outperforms other modeling alternatives, even with the more complex WOFOST model. Secondly, we showed that DMC-MTL effectively optimizes per-stage phenology predictions which are key to vineyard management operations. Finally, our results on parameter smoothing and weather windows showed that our assumptions for our DMC-MTL hybrid model were both appropriate and necessary to find the best fit model with the available data.

\end{document}